\documentclass[11pt,a4paper,extrafontsizes,oneside]{article}
\pdfoutput=1

\usepackage{authblk}
\usepackage[utf8]{inputenc}
\usepackage[T1]{fontenc}
\usepackage{hyperref}
\usepackage{url}
\usepackage{booktabs}
\usepackage{amsfonts}
\usepackage{nicefrac}
\usepackage{microtype}      
\usepackage{multirow}
\usepackage{float}
\usepackage{array}
\newcolumntype{L}{>{\centering\arraybackslash}m{2cm}}
\usepackage{subcaption}
\usepackage{graphicx}
\usepackage{color}
\usepackage{algorithmicx}
\usepackage{algorithm}
\usepackage{enumitem}
\usepackage{graphicx}
\usepackage[labelfont=bf]{caption}
\usepackage{amsmath}
\usepackage{wrapfig}

\begin{document}

\title{\Large A Novel Data-Driven Framework for Risk Characterization and
Prediction from Electronic Medical Records: A Case Study of Renal Failure}

\author[1]{Prithwish Chakraborty\thanks{prithwish.chakraborty@ibm.com}}
\author[1]{Vishrawas Gopalakrishnan\thanks{vishrawas.gopalakrishnan1@ibm.com}}
\author[3]{Sharon M.H. Alford\thanks{shensle@us.ibm.com}}
\author[4]{Faisal Farooq\thanks{ffarooq@us.ibm.com}}
\affil[1]{IBM Watson Health, MA, USA}
\affil[2]{IBM Watson Health, MI, USA}
\affil[3]{IBM Watson Health, NY, USA}
\date{}
\maketitle

\begin{abstract}

Electronic medical records (EMR) contain longitudinal information about
patients that can be used to analyze outcomes. 
Typically, studies on EMR data have worked with established variables that have
already been acknowledged to be associated with certain outcomes.
However, EMR data may also contain hitherto unrecognized factors 
for risk association and prediction of outcomes for a disease.
In this paper, we present a scalable data-driven framework to analyze
EMR data corpus in a disease agnostic way that systematically uncovers
important factors influencing outcomes in patients, \textit{as supported by data and 
without expert guidance}.
We validate the importance of such factors by using the framework to predict
for the relevant outcomes.
Specifically, we analyze EMR data covering approximately 47 million unique
patients to characterize renal failure (RF) among type 2 diabetic (T2DM)
patients.
We propose a specialized L1 regularized Cox Proportional Hazards (CoxPH) 
survival model to identify the important factors from those available from 
patient encounter history.
To validate the identified factors, we use a specialized generalized linear 
model (GLM) to predict the probability of renal failure for individual patients
within a specified time window.
Our experiments indicate that the factors identified via our data-driven method
overlap with the patient characteristics recognized by experts. Our
approach allows for scalable, repeatable and efficient utilization of data
available in EMRs, confirms prior medical knowledge and can generate new
hypothesis without expert supervision.
\end{abstract}

\section*{Introduction}
The growth and wide-spread adoption of Electronic Medical Record (EMR) systems
~\cite{charles2013adoption,wallace2014multi,hsiao2010electronic} has led to
efficient recording of patient characteristics and their encounter details that
makes the information accessible for secondary data purposes.  The resultant
rich longitudinal view of the patients’ disease progression
and treatment response, allows for novel analyses, even leading to revision of
the recommended treatment and patient-care
plans~\cite{jensen2012mining,masuda2002framework,ohno2015mining}.  Machine
learning approaches furthermore provide robust methods to harvest and compute
large data stores, alleviating past limitations of epidemiological studies on
clinical data~\cite{mani2012type}.  Increased adoption of EMR standards and
health information exchanges, primarily by online and cloud-based data
marts~\cite{chen2012secure}, have further led to efficient linking of patient
encounters/visits across hospitals and addresses many of the concomitant issues
with using EMR data such as  missing information and temporal blind-spots. Consequently, data driven clinical studies are now feasible 
and coupled with large EMR datasets can now, for the first time,
lead to discovery of hitherto unknown associations between complex chronic
diseases. 

In this work, we focus on a disease agnostic data-driven approach to identify
novel factors associated with disease characterization as supported by data and
without necessitating expert guidance.
Our framework is designed to be scalable and modifies 
the classic epidemiological survival model, Cox proportional hazard model~\cite{10.2307/2985181}.
The modifications allow us to identify relevant attributes and validate the same via outcome prediction
automatically. Specifically, we exploit the relationship between survival
model and Generalized Linear Model(GLM)~\cite{mccullagh1989generalized} to
create a L1 regularized proportional hazard model for
characterization backed by a L1 regularized GLM regression model to predict
the occurrence of adverse or critical events. This approach provides the
flexibility to overcome the traditional limitation of survival models,
specifically the tracking of the disease/event progression only up to the
maximum duration present in the training set. At the same time, survival
models provide a principled approach to handle ``right censored’' data
points that is data is truncated at the time of last follow-up without
knowing if the patient eventually developed the outcome of interest.
Compared to prior studies~\cite{mani2012type} our model by virtue of its
data-driven regularization can identify factors that are significant as
supported by the EMR instead of relying only on expert knowledge.
Newly revealed factors can then be verified from outcome prediction and be added to the list of early
warning indicators to be monitored. To illustrate our framework, we provide a
use-case characterizing renal failure (RF) in Type 2 Diabetes Mellitus (T2DM). 
To summarize, our key contributions are: 
\begin{itemize}[leftmargin=0.2in,noitemsep,nolistsep]
    
\item We present a scalable L1 regularized proportional hazard model to
  discover information from EMR about factors that can influence outcomes in
  patients, without expert guidance.

\item We present a GLM regression model to predict the
 occurrence of events amongst patients using factors discovered via survival
 modeling while adhering to the underlying generative process.

\end{itemize}

\section*{Methods}

\textbf{Experimental Setup:} We use a de-identified EMR corpus of
approximately $47$ million unique patients covering $1.6$ billion medical
encounters. We construct a type 2 diabetic cohort spanning the time period 
$1990$-$2016$ and covering $4$ million patients, following a three step 
definition: (a) patients must have been diagnosed with type 2 diabetes
anytime within this period (presence of snomed concepts), (b) they must have
had abnormal HbA1C levels (HbA1C $\geq 5.7$) from lab observations corresponding
to such diagnosis, and (c) had at least $3$ diabetes related encounters within 
this period.
A patient satisfying all the three criteria is considered to be diabetic and
included in the cohort and, the first time-point when these criteria are satisfied 
is considered to be the first diagnosis date for the patient.
This definition ensures that patients have been identified as diabetic from 
both procedural and diagnostic methods, thus reducing false positives.
From the EMR records, we create normalized medical histories of patients 
by resolving observations and outcomes with encounters. 
Furthermore, we admit both positive and negative cases of renal failure (acute and
chronic).  

\textbf{Feature Identification:} Typically, experts look at a fixed set of
factors such as age, weight and HbA1c level in blood while
analyzing renal failures in diabetic patients.
In this study, we aim to identify other possible indicators, as supported by 
the data, from the corpus of near-complete patient medical histories.
Following classical bio-statistical approaches, risk factors for 
extreme outcomes like renal failure can be characterized using
survival models such as Cox Proportional Hazard (CoxPH) models~\cite{10.2307/2985181}.
In these models, conditional event rates or hazards ($\lambda (t)$) associated
with factors $X_i = \{X_{i,1}, \dots, X_{i,p}\}$ characterizes the 
corresponding risk of outcomes which, for a patient $i$, can 
be given as:
\[
\lambda (t|X_{i})  =  \lambda _{0}(t)\exp(\beta _{1}X_{i1}+\cdots +\beta _{p}X_{ip}) = \lambda _{0}(t)\exp(X_{i}\cdot \beta )
\]
where, $\lambda_{0}(t)$ denotes the baseline hazard at time $t$ (independent of
patient covariates) and $\beta_{j}$ 
denotes the regression weight of the factor $j$. 
To identify and characterize such factors from the data, we use 
L$1$ regularized survival models which have been proven to be effective in 
other domains~\cite{penalized}. Specifically, we find these parameters by fitting
the regularized log-likelihood ratio (eq.~\ref{eq:coxPH_lasso}) against 
normalized EMR history of the patients in the cohort.

\vspace{-1em}
\begin{equation}
\label{eq:coxPH_lasso}
  \ell (\beta ) =  \sum _{j}(\sum _{i\in H_{j}}X_{i}\cdot \beta - \sum _{\ell =0}^{m-1}\log (\sum _{i:Y_{i}\geq t_{j}}\theta _{i}-{\frac {\ell }{m}}\sum _{i\in H_{j}}\theta _{i})) +\gamma \|\beta \|_{1}
\end{equation}
where, $\gamma$ is the L$1$ penalty factor and we identify the factors which 
are (a) statistically significant for survival outcomes and (b) associated 
with the largest regression weights ($\beta ( > 0)$).

\textbf{Renal Failure Prediction:} 
We validate the data-driven factors by using these as covariates to predict for 
renal failure in type 2 diabetic patients based on their medical histories.
Although survival models are not particularly suited for prediction on
time-intervals outside the training data,
we note that the partial log-likelihood in CoxPH models have close ties to 
logistic GLM regressions~\cite{mccullagh1989generalized}.
We use a regularized logistic regression (eq.~\ref{eq:logistic_lasso}) 
to use these factors to predict for the possibility of renal failure.
We use the average history of factors within a pre-defined observation period (first $n \in \{$6$, $9$,
$12$\}$ months from first diagnosis date) in a rolling window setup 
and predict the occurrence of renal failure in the prediction period 
(the next $3$ months). 
\begin{equation}
  \label{eq:logistic_lasso}
  \ell_{logistic} = \underset{w,d}{min\,} \|w\|_1 + D \sum{}_{i=1}^n \log(\exp(- y_i (X_i^T w + d)) + 1)
\end{equation} 
\vspace{-3em}

\section*{Results}
% \begin{wraptable}{r}{0.63\textwidth}
\begin{table}
  \centering
  \scriptsize
  \caption{Top data-driven factors for renal failure in T2DM patients
  as extracted from EMR.}
  \label{tab:tb_surv_lasso:ranking}
\small
\begin{tabular}{|l|r|r|}
\hline
          LOINC  & description & coef. value \\
\hline
\hline
 8277-6    & Body surface Area &  1.536450e-01 \\
 28542-9   & Platelet mean volume &  2.762536e-02 \\
 30180-4   & Basophils/100 leukocytes  & 2.447092e-02 \\
 ET3111-5  &  & 1.967411e-02 \\
 6299-2    & Urea Nitrogen &  1.112565e-02 \\
 26450-7   &  Eosinophils/100 leukocytes &  8.770401e-03 \\
 3094-0    &  Urea nitrogen [Mass/volume] &3.679680e-03 \\
 ET3123-9  &   &3.000471e-03 \\
 26511-6   & Neutrophils/100 leukocytes & 2.930090e-03 \\
 26444-0   & Basophils [\#/volume] in Blood & 1.513770e-03 \\
 2345-7    & Glucose & 5.521050e-04 \\
 26484-6   & Monocytes  & 4.750066e-04 \\
 8480-6    & Intravascular systolic & 2.682153e-04 \\
 2339-0    & Glucose [Mass/volume] in blood & 1.919959e-04 \\
 2571-8    & Triglyceride & 1.305691e-04 \\
 8310-5    & Body temperature & 1.291046e-04 \\
 1920-8    & Aspartate aminotransferase & 1.268876e-04 \\
 1742-6    & Alanine aminotransferase & 1.218098e-04 \\
 26499-4   & Neutrophils & 2.527398e-05 \\
 14957-5   & Albumin & 8.445566e-07 \\
 8867-4    & Heart Rate & 9.496403e-13 \\
\hline
\end{tabular}
\end{table}
% \end{wraptable}
Table~\ref{tab:tb_surv_lasso:ranking} tabulates the top statistically significant
data-driven factors w.r.t. survival as 
identified via eq.~\ref{eq:coxPH_lasso} from our T2DM patient
corpus. Table~\ref{tab:tb_surv_lasso:ranking} shows that our method uncovers several observational attributes,
\textit{from the data} and \textit{without expert input}, that are 
potentially correlated with an increased risk of renal failures.
We validate the importance of these factors using our predictive model as shown
in Table~\ref{tab:tb_prediction_acc} which shows that the outcome prediction
performance increases when expert factors are supplemented by data-driven
factors - thus indicating that the data-driven factors may be uncovering hidden
correlations and thus, of potential interest to experts.
Here, we found the L$1$ penalty using $5$ fold cross validation and 
for each observation period, we used $10$ fold cross-validation to quantify
accuracy, sensitivity, and specificity.

\begin{table}[!hb]
  \centering
  \scriptsize
  \caption{Comparison of outcome prediction performance for renal failure
  amongst Type 2 Diabetic patients.  Combination of data-driven and expert
  factors exhibits the  best overall performance.}
  \vspace{0.5em}
  \label{tab:tb_prediction_acc}
\scriptsize
\begin{tabular}{|l|r|r|r|}                   
  \hline                                      
     Model attributes             &  Accuracy &  Specificity & Sensitivity \\
  \hline                                        
  \hline                                      
Only expert features              &  0.86     & 0.89 & 0.68 \\
Data driven features              &  0.89     & 0.92 & 0.71 \\
expert + data -driven features    &  0.91     & 0.93 & 0.73 \\
\hline
\end{tabular}
  \end{table}

\section*{Discussion}

\begin{figure*}[bt]
    \centering
    \begin{subfigure}[t]{0.32\textwidth}
        \centering
        \includegraphics[width=0.9\linewidth]{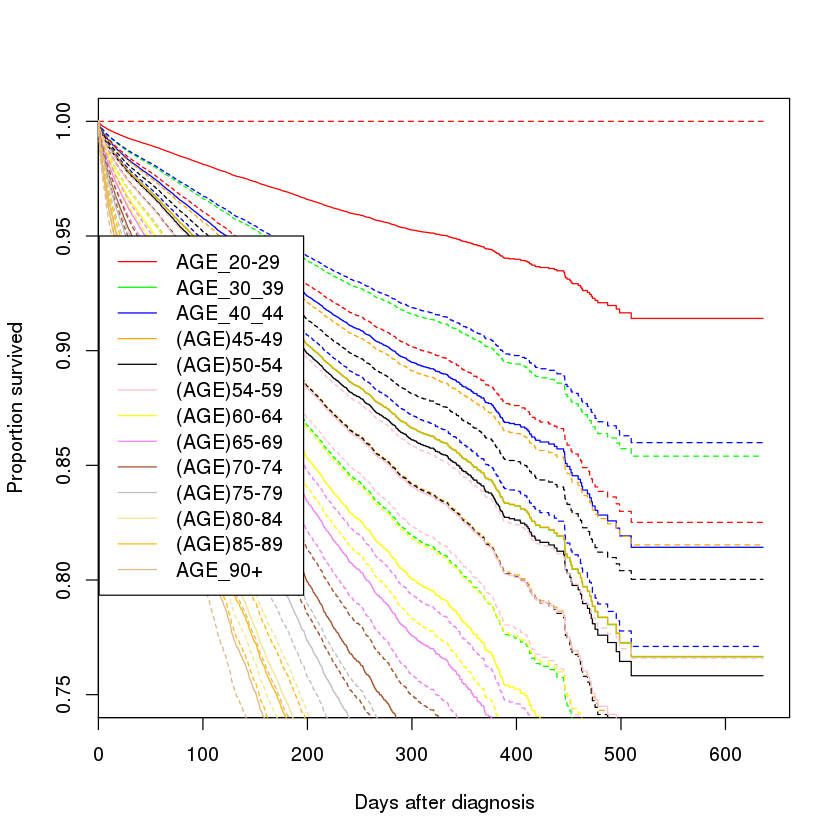}
        \caption{Age}
    \end{subfigure}    \hspace{-1em}
    \begin{subfigure}[t]{0.32\textwidth}
        \centering
        \includegraphics[width=0.9\linewidth]{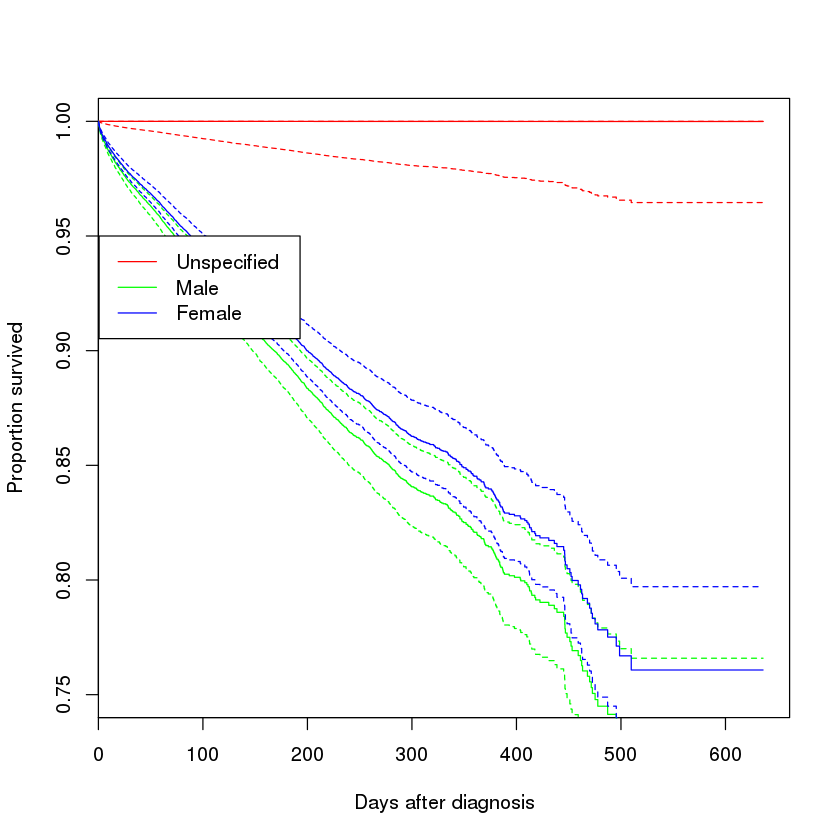}
        \caption{Gender}
    \end{subfigure}    \hspace{-1em}
    \begin{subfigure}[t]{0.32\textwidth}
        \centering
        \includegraphics[width=0.9\linewidth]{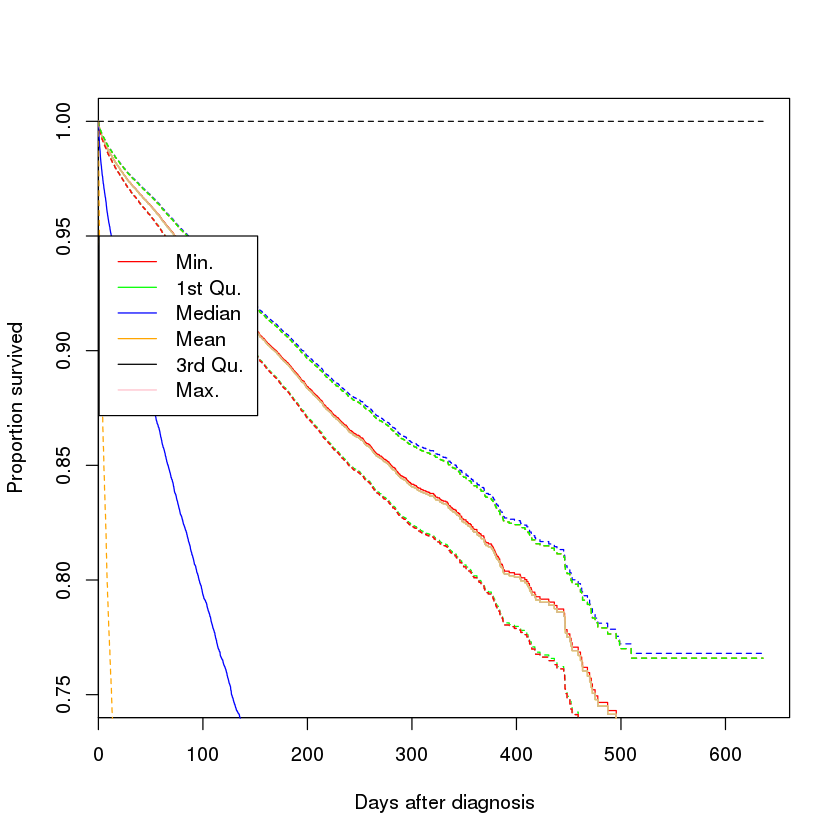}
        \caption{HbA1c}
    \end{subfigure}    \\
    \begin{subfigure}[t]{0.32\textwidth}
        \centering
        \includegraphics[width=0.9\linewidth]{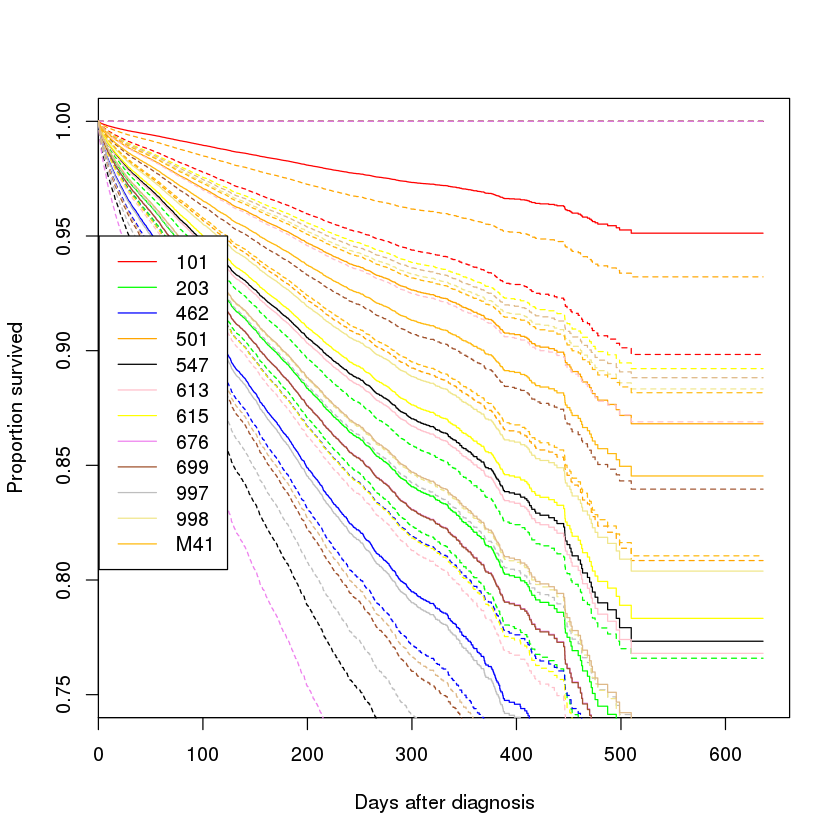}
        \caption{Race}
    \end{subfigure}
    \hspace{-1em}
    \begin{subfigure}[t]{0.32\textwidth}
        \centering
        \includegraphics[width=0.9\linewidth]{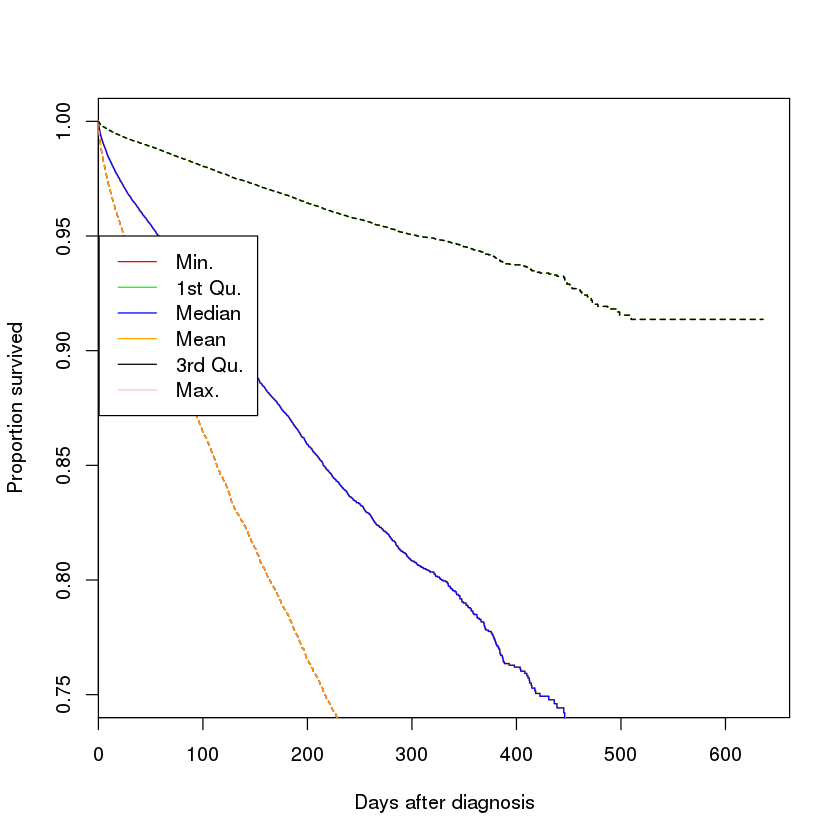}
        \caption{Weight}
    \end{subfigure}    \hspace{-1em}
    \begin{subfigure}[t]{0.32\textwidth}
        \centering
        \includegraphics[width=0.9\linewidth]{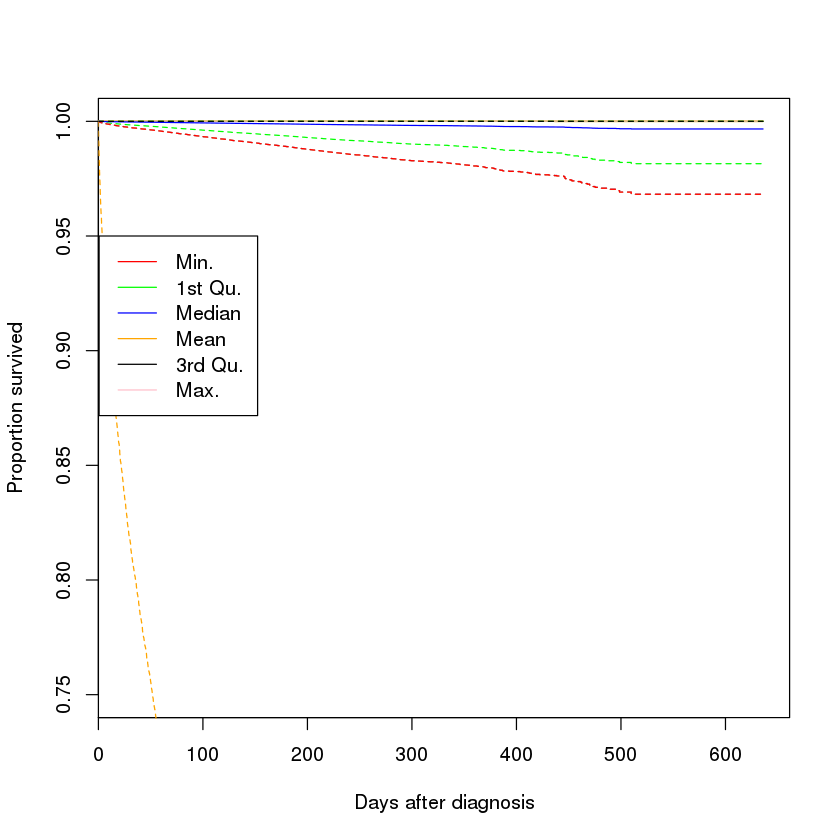}
        \caption{LOINC 8227-6}
    \end{subfigure}
    \caption{Adjusted baseline survival ratio with respect to individual 
    factors}
    \label{fig:adjusted_survival}
\end{figure*}

In this section we briefly discuss some interesting findings from
our assay.

\textbf{Relative importance of factors :}
We ascertain the relative importance of factors using coefficient
rankings in Table~\ref{tab:tb_surv_lasso:ranking},
where the coefficients are calculated in the presence of other factors.
Following classical survival analysis methodology, we further investigate the importance of
each individual factor using adjusted survival ratios. We create the adjusted
ratio for a factor by marginalizing against other factors, setting such factors to their 
mean or most prevalent value (depending on the type of the factor).
Figure~\ref{fig:adjusted_survival} shows the adjusted ratios for the most important
data driven as well as Expert factors. As can be seen when marginalized, we can
uncover certain characteristics within the factors. For example, males are generally
at higher risk than females. While weight and renal risk exhibits an inverse ratio,
HbA1c shows a direct ratio with renal failure risk. Interestingly, the difference
is not as marked in the case of LOINC 8277-6.

\textbf{Are there any significant factors not identified by survival models :} 
Although weight is one of the factors considered by experts, 
it ranks low from our data-driven framework. We posit that this could indicate that
that effects identified by weight may already be covered by other factors and
hence its ranking is not significant. Interestingly,
the importance of gender and race remain constant throughout different models - 
indicating that these are fundamental factors. A possible extension to analyze 
these factors in depth could be a stratified approach. However, such approaches
can obscure the importance of the fundamental factors

\begin{figure}
% \begin{wrapfigure}{r}{0.37\textwidth}
  \centering
  \includegraphics[width=0.95\linewidth]{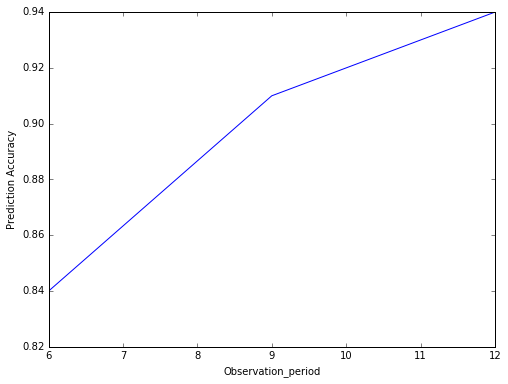}
  \caption{Prediction accuracy as a function of observation period}
  \label{fig:pred_trace}
% \end{wrapfigure}
\end{figure}
\textbf{How does the observation period affect the prediction performance :}
Table~\ref{tab:tb_prediction_acc} reports the overall prediction accuracy for
our experimental setup. However, the prediction performance when viewed as a
function of the observation period (see Figure~\ref{fig:pred_trace}), uncovers
interesting patterns.  It can be seen that, in general, increasing the
observation period increases the prediction accuracy.
It is to be noted, increasing the observation doesn't necessarily increase the
amount of data observed in general (as we are taking average history of
patients who haven't had a renal failure in the observation period).  One
possible explanation could be the fact that with the increase in observation
period, more renal failures happen in the subsequent prediction period and as
such the prediction problem becomes less skewed and well-defined. This
indicates that while dealing with low observation periods we may need to
include additional strategies such as SMOTE~\cite{chawla2002smote}.
\vspace{-1em}

\section*{Conclusion}
\vspace{-1em}
In this paper, we propose a systematic framework for identifying risk factors
for outcomes as supported by data from EMR corpus. We showcase our framework on
renal failures for diabetic patients and discussed various components of the
same. Our experiments indicate that the framework is able to identify
significant factors.  However, certain factors as identified by experts
don't rank highly from this framework. Further research is needed to model hidden interactions between
factors to distinguish between `fundamental' factor and derivative factors.

% \clearpage

\end{document}